\documentclass[11pt,a4paper]{article}
\usepackage[hyperref]{emnlp2020}

\usepackage[utf8]{inputenc} % allow utf-8 input
\usepackage{microtype}
\usepackage{graphicx}
\usepackage{subfigure}
\usepackage{booktabs} 
\usepackage{comment}
\usepackage{colortbl}
\usepackage{arydshln}
\usepackage{makecell}
\usepackage{hyperref}
\usepackage{mathtools}
\usepackage{color}
\usepackage{graphicx}
\usepackage{CJKutf8}
\usepackage{url}
\usepackage{multicol,multirow}
\usepackage{makecell}
\usepackage{array}
\usepackage{bm}
\usepackage{parskip}
\usepackage[T1]{fontenc}    % use 8-bit T1 fonts
\usepackage{booktabs}       % professional-quality tables
\usepackage{amsmath, amsthm}
\usepackage{amssymb}
\usepackage{amsfonts}

\usepackage{times}
\usepackage{latexsym}

\usepackage{microtype}
\definecolor{ao}{rgb}{0.0, 0.5, 0.0}
\definecolor{asparagus}{rgb}{0.53, 0.66, 0.42}
\definecolor{amber}{rgb}{1.0, 0.49, 0.0}
\definecolor{alizarin}{rgb}{0.82, 0.1, 0.26}
\definecolor{applegreen}{rgb}{0.55, 0.71, 0.0}
\definecolor{amethyst}{rgb}{0.6, 0.4, 0.8}
\definecolor{auburn}{rgb}{0.43, 0.21, 0.1}
\aclfinalcopy % Uncomment this line for the final submission
\title{Self-Explaining Structures Improve NLP Models}

\date{}

\author{
Zijun Sun$^\clubsuit$, Chun Fan$^{\spadesuit\bigstar}$, Qinghong Han$^\clubsuit$, Xiaofei Sun$^\clubsuit$, \\ 
{\bf Yuxian Meng$^\clubsuit$, Fei Wu$^\blacklozenge$ and Jiwei Li $^{\blacklozenge\clubsuit}$}\\
  $^\blacklozenge$Zhejiang University,
  $^\spadesuit$Computer Center of Peking University \\
  $^\bigstar$Peng Cheng Laboratory,  
  $^\clubsuit$ Shannon.AI\\
  \{zijun\_sun, jiwei\_li\}@shannonai.com,
  fanchun@pku.edu.cn,
  wufei@zju.edu.cn
}

\begin{document}
\maketitle

\begin{abstract}
Existing approaches to explaining deep learning models in NLP usually suffer from two major drawbacks:
(1) the main model and the explaining model are decoupled: 
an additional probing or surrogate model is used to  interpret an existing model, and  thus 
existing explaining tools
 are not self-explainable; (2) the probing model is only able to explain a model's predictions by operating on 
 low-level features  
by computing  saliency scores for individual words 
but  are clumsy  at  high-level text units such as phrases, sentences, or paragraphs. 
%, such as phrase, sentence or paragraph due to the entanglements and highly non-linear operations in semantic compositions. 

To deal with these two issues, in this paper, we propose a 
simple yet general and effective 
self-explaining framework  for deep learning models in NLP. 
The key point of the proposed framework is to put an additional layer, as is called by the {\it interpretation layer},  on top of any existing NLP model.  
This layer aggregates the information for each text span, which is then associated with a specific weight, and their weighted combination is fed to the softmax function for the final prediction. 

The proposed model comes with the following merits: 
(1) span weights make the model  self-explainable and do not require an additional probing model for interpretation; 
(2) the proposed model
 is general and can be adapted to any existing deep learning structures in NLP; 
(3) the weight associated with each text span 
 provides direct importance 
  scores for higher-level text units such as phrases and sentences.
We for the first time show that interpretability does not come at the cost of performance: 
a neural model of self-explaining features obtains better performances than its counterpart without the 
self-explaining nature, 
 achieving a new SOTA performance of {\bf 59.1} on SST-5 and 
  a new SOTA performance of {\bf 92.3} 
 on SNLI. \footnote{Code is available at \url{https://github.com/ShannonAI/Self_Explaining_Structures_Improve_NLP_Models}}
%We wish this work would foster more researches in developing NLP models that offer  high performance and interpretability at the same time.
\end{abstract}

%%%%%%%%%%%%%%%%%%%%%%%%%%%%%%%
% start a new section
%%%%%%%%%%%%%%%%%%%%%%%%%%%%%%%
\section{Introduction} 
A long-term criticism against deep learning models is the lack of interpretability \cite{simonyan2013deep,bach2015lrp,montavon2017explaining,kindermans2017learning}. 
The black-box nature of neural models not only significantly limits the scope of applications of deep learning models (e.g., in  biomedical or legal domains), 
but also hinders model behavior analysis and error analysis. 

To enhance neural models' interpretability, various approaches to rationalize neural models' predictions have been proposed (details see Section \ref{related-work}). 
%e.g., \cite{simonyan2013deep,hermans2013training,datta2016algorithmic,greff2016lstm,ribeiro2016i,koh2017understanding,shrikumar2017learning, kadar2017representation,adler2018auditing,tenney2019bert,clark2019does,wallace2019allennlp,han2020explaining}. 
%Among these methods, one of the most widely-adopted strategy is the saliency approach \cite{simonyan2013deep,li2015visualizing,li2016understanding,selvaraju2017grad,meng2020pair}, which explains neural network predictions by identifying the salient word(s) within the input that the model attends to. The saliency score can be obtained by calculating  first-order gradients \cite{simonyan2013deep,li2015visualizing,selvaraju2017grad} or measuring the impact of erasing or perturbing a certain input word \cite{li2016understanding,feng2018pathologies}. 
Existing interpretation models have two major drawbacks. Firstly, they are not self-explainable and require 
a  
 probing model  to be additionally built to interpret the original model. 
 Building the probing model is not only an additional burden, but more importantly, 
 the intrinsic decoupling nature of the two models makes  
 the probing model only able to provide approximate interpretation results. 
 Take the gradient-based saliency model \cite{simonyan2013deep,li2015visualizing,selvaraju2017grad} as an example, the  derivative of the target probability (or the logit) w.r.t. the input dimensions is only 
 an approximate feature weight 
  in the
  first-order Taylor expansion, with all higher  order items omitted. 

Secondly, interpretation models mostly focus on
learning 
 word-level importance scores assigned by the  the main model 
  and are hard to be adapted to higher-level text units such as phrases, sentences, or paragraphs. A straightforward way to compute the saliency score for a phrase or a sentence could be averaging the scores for its constituent words. 
Unfortunately, this over-simplified strategy is inadequate to capture the 
 semantic composition in language, which  
  involves multiple layers of highly non-linear operations in neural nets, and can thus result in high risk of misinterpretation. 
  As pointed by \newcite{murdoch2018beyond}, interpreting neural net predictions should go beyond word level. 

We raise the following question: {\bf what makes a good interpretation method for NLP?}
Firstly,  the method should be self-explainable and no additional probing model is needed.
Secondly, the model should offer precise and clear saliency scores for any level of text units. 
Thirdly, the self-explainable nature does not tradeoff performances.

Towards these three purposes, in this paper, we propose a self-explainable framework for deep neural models in the context of NLP. 
The key point of the proposed framework is to put an additional layer, as is called the {\it interpretation layer} on top of any existing NLP model,  
and this layer aggregates the information for all  (i.e., $\mathcal{O}(n^2)$) text spans. 
Each text span is  associated with a specific weight, and their weighted combination is fed to the softmax function for the final prediction. 
The proposed 
 structure offers  following advantages:
(1) the model is self-explainable: 
the interpretation layer is trained along with the objective, with no  probing model needed. 
 Weights at the interpretation layer can directly be used as saliency scores for  corresponding text spans; 
(2) since the saliency score for any text span can be straightforwardly derived from the interpretation layer,  
the model offers direct,  precise and clear saliency scores for any  level of text units beyond word level. 
 (3) the interpretation layer collects information for each text span in the form of representations, and forwards the weighted sum to the final prediction layer. 
Therefore, no additional information is incorporated, nor any information is lost at the interpretation layer. This makes the model not come at the cost of performances. 

The proposed framework is general, and can be adapted to any prevalent deep learning structure in NLP.
We show that the proposed framework (1) offers clear model interpretations; 
(2) facilitates error analysis; 
(3) can help   downstream NLP tasks such as adversarial example generation; and 
(4) most importantly, for the first time proves that self-explainable nature does not come at the cost of performances, but rather, 
leads to better results in NLP,  
 achieving a new SOTA performance of {\bf 59.1} on SST-5 and 
  a new SOTA performance of {\bf 92.3} 
 on SNLI.

\begin{comment}
The rest of this work is organized as follows: related work is detailed in Section 2.
The proposed self-explaining framework is illustrated in Section 3.
Evaluation methods and experimental results are respectively shown in Section 4 and 5, followed by a brief conclusion in Section 6. 
\end{comment}

\section{Related Work} 
\label{related-work}
Rationalizing model predictions is of growing interest \citep{ribeiro2016i,shrikumar2017learning,10.1371/journal.pone.0130140,kindermans2017learning,montavon2017explaining,schwab2019cxplain,arrieta2020explainable}.
%\citet{ribeiro2016i} proposed LIME to approximate  model behaviors  using a surrogate model over perturbed inputs. \citet{10.1371/journal.pone.0130140} used Layer-Wise Relevance Propagation (LRP), a technique that progressively aggregates relevance scores of each pixel from higher layers to lower layers, to explain which pixels of an image are relevant to a model's decision.
In NLP, 
approaches to interpret neural models include 
 extracting 
pieces of input text, called ``rationales'', as justifications to model predictions  \citep{lei2016rationalizing,chang2019game,deyoung-etal-2020-eraser,jain-etal-2020-learning}, studying the 
efficacy and dynamics of hidden states in recurrent networks \citep{karpathy2015visualizing,shi-etal-2016-neural,greff2016lstm,strobelt2016visual}, and applying variants of the attention mechanism \citep{bahdanau2014neural} to interpret model behaviors \cite{jain2019attention,serrano2019attention,wiegreffe-pinter-2019-attention,vashishth2019attention,rogers2020primer}. 
\citet{lei2016rationalizing} proposed to extract text snippets as model explanations.
%To extract proper snippets, another prediction model is trained to take the extracted snippet as input to maximize  the probability of  output labels. 
\citet{rajani2019explain} collected human rationales for commonsense reasoning.
% on which a model is trained to generate open-ended explanations to improve performances on various downstream tasks.
Other works \citep{chen-etal-2019-seeing,lehman-etal-2019-inferring,chai2020description} trained independent models to extract supporting sentences as auxiliary guidelines for 
downstream tasks.
Using attentions as a tool for model interpretation, \citet{ghaeini2018interpreting} visualized attention heatmaps to understand how natural language inference models build interactions between two sentences;
\citet{vig2019analyzing,tenney2019bert,clark2019does,htut2019attention} analyzed the attention structures by plotting heatmaps, and found that meaningful linguistic patterns
exist in different heads and layers.
Despite the interpretability the attention mechanism offers, 
\newcite{serrano2019attention,jain2019attention} observed the highly inconsistency between attentions and predictors, and suggested that attentions
should not be treated as justification for a decision.

Saliency methods are widely used in  computer vision \citep{simonyan2013deep,zeiler2014visualizing,springenberg2014striving,adler2018auditing,datta2016algorithmic,srinivas2019full} and NLP \cite{denil2015extraction,li2015visualizing,li2016understanding,arras2016explaining,ebrahimi2017hotflip,feng2018pathologies,meng2020pair} for model interpretation. 
The key idea is to 
 find the  salient features responsible for a model's prediction.
 \citet{simonyan2013deep,srinivas2019full} visualized the contributions of input pixels by compute the derivatives of the label logit in the output layer with respect to the input pixel. 
 \citet{adler2018auditing,datta2016algorithmic} explained neural models by perturbing different  parts of the input, and compared the performance change in downstream tasks to measure the importance of perturbed features.
In the context of NLP,  \citet{denil2015extraction} used saliency maps to identify and extract task-specific salient sentences from documents to maximally preserve document topics and semantics; 
\citet{li2015visualizing} visualized word-level saliency maps to understand how individual words affect model predictions; \citet{ebrahimi2017hotflip} crafted white-box adversarial examples 
to find the most salient text-editing operations (flip, insertion and deletion) to trick models
by computing derivatives w.r.t. these editing operations; \citet{meng2020pair} combined the saliency method and the influence function \citep{koh2017understanding} to interpret model predictions from both learning history and inputs.

Our work is inspired by \newcite{melis2018towards}, which achieves  ``self-explaining'' by jointly training a deep learning model and a linear regression model with human-interpretable features.
%, and ensuring that  the deep learning model  behaves similar to the linear regression model. 
It is worth noting that the model in \newcite{melis2018towards} still requires a surrogate model, i.e., the  linear regression model with human-interpretable features. 
Instead, the proposed framework does not require a surrogate model. 
Our work is also inspired by \newcite{selvaraju2017grad} in computer vision, 
%Instead of computing gradients w.r.t. individual pixels of an image, \newcite{selvaraju2017grad} proposes to
which 
computes the gradient  w.r.t.  feature maps of the high-level convolutional
layer to obtain high-level saliency scores. 

\section{The Self-Explaining Framework}
\subsection{Notations}
Given an input sequence $\bm{x} = \{x_1, x_2, ..., x_N\}$, where $N$ denotes the length of $\bm{x}$. 
Let $\bm{x}(i,j)$ denote the text span starting at index $i$, ending at index $j$, where
$\bm{x}(i,j) = \{x_i, x_{i+1},..., x_{j-1}, x_j\}$. 
We wish to predict the label $y$ for $\bm{x}$ based on $p(y|\bm{x})$.
Let $V$ denote vocabulary size, and word representations are stored in $\bm{W}\in\mathbb{R}^{ V\times D}$.
The input $\bm{x}$ is associated with the label $y\in \mathcal{Y}$. 
\subsection{Model}
\begin{figure}
    \centering
    \includegraphics[scale=0.6]{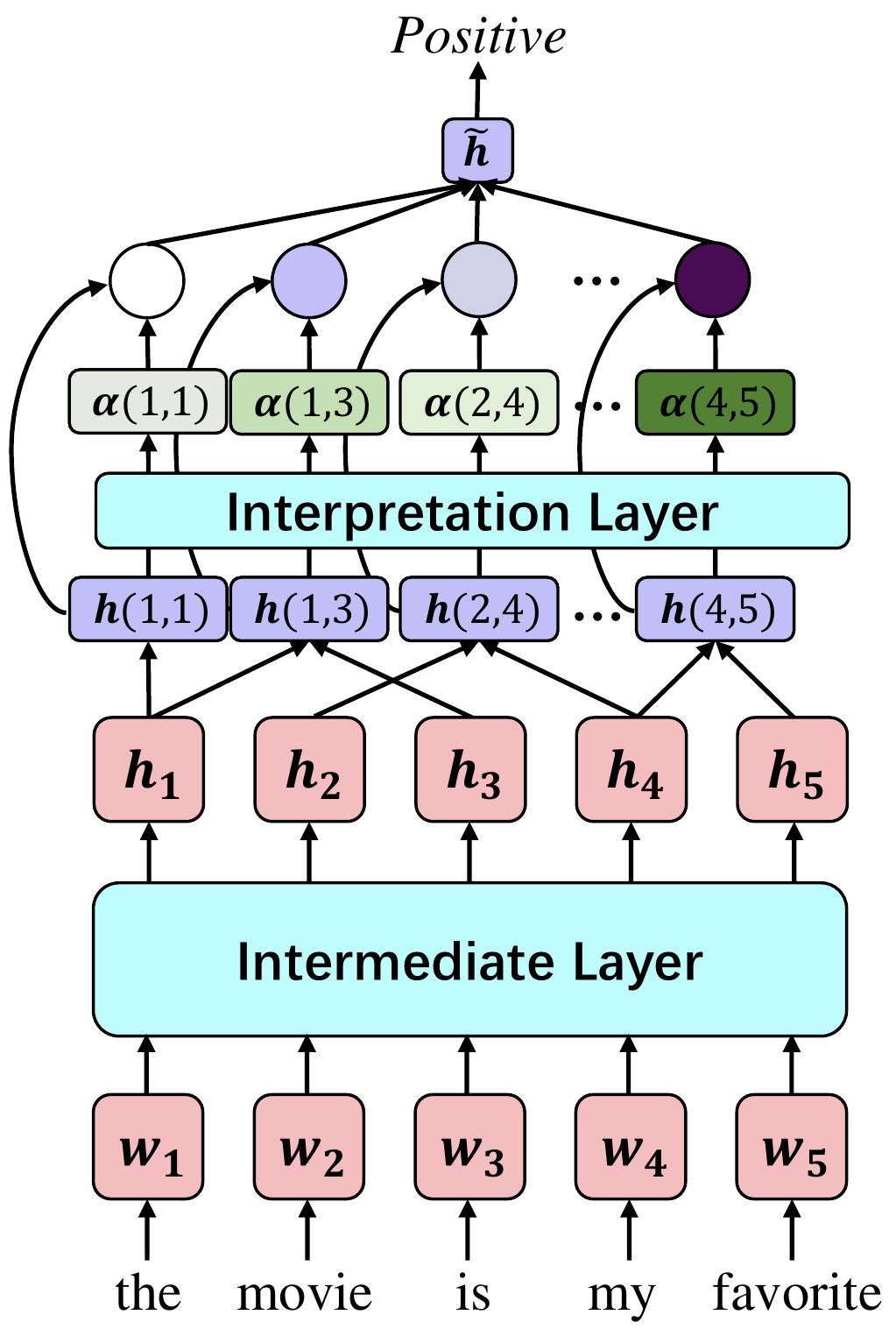}
    \caption{An overview of the proposed model.}
    \label{fig:overview}
\end{figure}

For illustration purposes, we use transformers \cite{vaswani2017attention} as the backbone to show how the proposed framework works. The framework can be easily extended to other models such as BiLSTMs or CNNs. An overview of the proposed model is shown in Figure \ref{fig:overview}.

{\bf Input Layer} Similar to standard deep learning models in NLP, the input layer for the proposed model consists of the stack of word representations for words in the input sequence, where $x_t$ is represented as a $D$-dimensional vector $\bm{W}[x_t, :]$. 

{\bf Intermediate Layer}
On top of the input layer are the $K$ encoder stacks, where each stack consists of multi-head attentions, layer normalization and residual connections. 
The representation for $k$-th layer at position $t$ is denoted by $\bm{h}_t^k \in \mathbb{R}^{1\times D}$. 
Specially, the presentation for the last layer at position $t$ is denoted by $\bm{h}_t^K \in \mathbb{R}^{1\times D}$. 

{\bf Span Infor Collecting Layer (SIC)}
To enable direct measurement for saliency of an arbitrary text span, we place a Span Infor Collecting Layer on top of intermediate layer. 
For an arbitrary text span $\bm{x}(i,j)$, we first obtain a dense representation $\bm{h}(i,j)$ to represent $\bm{x}(i,j)$, and $\bm{h}(i,j)$ needs to contain all
semantic and syntactic 
 information stored in $\bm{x}(i,j)$. 
Concretely, $\bm{h}(i,j)$ is obtained by taking the representation for the starting index at the last intermediate layer $\bm{h}^K_i$, and the representation for the ending index $\bm{h}^K_j$:
 \begin{equation}
  \bm{h}(i,j) = F(\bm{h}^K_i, \bm{h}^K_j)
 \label{combine}
 \end{equation}
where $F$ denotes the mapping function, which be FFN or other forms.
The details of $F$ will be discussed in section \ref{F}. The strategy of using the starting and ending representations to represent a text span has been used in many recent works on  span-level features in texts \cite{li2019unified,joshi2020spanbert,wu2019coreference}.
The SIC layer iterates over all text spans, and collects $\bm{h}(i,j)$ for all $i\in [1,N]$, $j\in [i,N]$, with the complexity being $\mathcal{O}(N^2)$.

{\bf Interpretation Layer} The interpretation layer aggregates information from all text spans $\bm{h}(i,j)$:
this is achieved by first 
 assigning weights $\alpha(i,j)$ to each span $\bm{x}(i,j)$ and combing these representations using weighted sum. The weight $\alpha(i,j)$ 
 can be obtained by first mapping $\bm{h}(i,j)$ to a scalar, and then normalizing all $\alpha(i,j)$: 
\begin{equation}
\begin{aligned}
o(i,j)& = \hat{\bm{h}}^\top \bm{h}(i,j) \\
\alpha(i,j)&= \frac{\exp(o(i,j))}{\sum_{i\in [1,N], j\in [i,N]} \exp(o(i,j))}  
\end{aligned}
\end{equation}
where $\hat{\bm{h}}\in \mathbb{R}^{1\times D}$. The output $\tilde{\bm{h}}\in \mathbb{R}^{1\times D}$ from the  interpretation layer is the weighted average of all span representations:
\begin{equation}
  \tilde{\bm{h}} = \sum_{i\in [1,N], j\in [i,N]} \alpha(i,j) \bm{h}(i,j)
\end{equation}
{\bf Output Layer} Similar to the standard setup, the output layer of the proposed framework is the probability distribution over labels using the softmax function:
\begin{equation}
\begin{aligned}
p(y|\bm{x}) = \frac{\bm{u}_y^\top \tilde{\bm{h}}}{\sum_{y\in \mathcal{Y}}\bm{u}_y^\top \tilde{\bm{h}} }
\end{aligned}
\end{equation}
where $\bm{u}_y\in\mathbb{R}^{D\times 1}$. 
As can be seen, $\alpha(i,j)$ measures how much $\bm{x}(i,j)$ contributes to the final representation $\tilde{\bm{h}}$, and thus indicates 
 the importance of the text span $\bm{x}(i,j)$. The proposed strategy is similar to gradient-based interpretation methods \cite{simonyan2013deep,li2015visualizing}. Using the chain rule, we have 
 \begin{equation}
 \begin{aligned}
&\frac{\partial p(y|\bm{x})}{\partial \bm{h}(i,j)}  = \frac{\partial p(y|\bm{x})}{\partial \tilde{\bm{h}}} \frac{\partial{\tilde{\bm{h}}}}{\partial {\bm{h}(i,j)}} \\
& =  \frac{\partial p(y|\bm{x})}{\partial \tilde{\bm{h}}} \left[\alpha(i,j)+ \bm{h}(i,j)\frac{\partial{\alpha(i,j)}}{\partial \bm{h}(i,j)}\right]
\end{aligned}
 \end{equation}
Omitting the second part, ${\partial p(y|\bm{x})}/{\partial \tilde{\bm{h}}}$  is approximately in proportion to 
$\alpha(i,j)$.
There is also a key advantage of the proposed framework over existing gradient-based interpretation methods, where 
the interpretation layer allows 
gradients to be straightforwardly computed with respect to an arbitrary text span. This is not feasible for vanilla gradient-based methods: 
because of the highly entanglement of neural networks, it's impossible to filter out the information of a specific text span 
from intermediate layers. Only gradients w.r.t. the input layer offers non-disentangling saliency scores, making the model only able to perform word-level explanations. 
\subsection{Efficient Computing}
\label{F}
A practical issue with Eq.\ref{combine} is the computation cost. If $F$ takes the form of FFN, the computational complexity for one single text span is $\mathcal{O}(D^2)$, 
leading to a final complexity of $\mathcal{O}(N^2D^2)$ if all spans are iterated over. This  is   computationally unaffordable for long texts. 
Towards efficient computations, we propose that $F$ takes the following form:
\begin{equation}
F(\bm{h}_i, \bm{h}_j) = \tanh[ \bm{W}(\bm{h}_i, \bm{h}_j, \bm{h}_i - \bm{h}_j, \bm{h}_i \otimes \bm{h}_j)]
\label{eff}
\end{equation}
where $\bm{W}= [\bm{W}_1, \bm{W}_2, \bm{W}_3, \bm{W}_4]$, $\bm{W}_i\in\mathbb{R}^{D\times D}$. $\otimes$ denotes the pairwise dot between two vectors. Elements in 
$(\bm{h}_i, \bm{h}_j, \bm{h}_i - \bm{h}_j, \bm{h}_i \otimes \bm{h}_j)$ respectively captures concatenation, element-wise difference and element-wise closeness between the two vectors, the strategy of which is been used in recent work to model interactions between two semantic representations \cite{mou2015natural,seo2016bidirectional}.

In Eq.\ref{eff}, $\bm{W}_1 \bm{h}_i$, $\bm{W}_2 \bm{h}_j$, $\bm{W}_3 \bm{h}_i$ and $\bm{W}_3 \bm{h}_j$ can be computed in advance for all $i$ and $j$, leading to a computational complexity of $\mathcal{O}(ND)$. 
For $\bm{W}_4 \bm{h}_i \otimes \bm{h}_j$, it can be factorized as $\sqrt{\bm{W}_4}\bm{h}_i \otimes \sqrt{\bm{W}_4}\bm{h}_j$, with each of the two parts being computed in advance. 
 The final computational complexity of $\bm{W}(\bm{h}_i, \bm{h}_j, \bm{h}_i - \bm{h}_j, \bm{h}_i \otimes \bm{h}_j)$ is thus $\mathcal{O}(ND)$. The element-wise tanh operation for all $N^2$ spans leads to a cost of $\mathcal{O}(N^2D)$, giving a complexity of $\mathcal{O}(N^2D)+ \mathcal{O}(ND) =\mathcal{O}(N^2D)$ for  the SIC layer, which significantly cuts the cost.
 The computational cost for the interpretation layer, which requires the dot product between $\hat{\bm{h}}$ and all $\bm{h}(i,j)$, is also $\mathcal{O}(N^2D)$,
 leading to the final  computational complexity $\mathcal{O}(N^2D)$. 
In this way, we reduce the computation cost from $\mathcal{O}(N^2D^2)$ to $\mathcal{O}(N^2D)$. 

\subsection{Training}
The training objective is the standard cross entropy loss. An additional regularization on $\bm{\alpha}$ is needed:  
the model should only focus on 
 a very small number of text spans. We thus wish the distribution of $\bm{\alpha}$ to be sharp. 
  We thus propose the following training objective:
 \begin{equation}
 \mathcal{L} = \log p(y|\bm{x}) + \lambda\sum_{i,j} \alpha^2(i,j)
 \label{objective}
 \end{equation}
which uses $\sum_{i,j} \alpha^2(i,j)$ as the regularizer. Under the constraint of $\sum_{i,j} \alpha(i,j)=1$, $\sum_{i,j} \alpha^2(i,j)$ achieves the highest value with one element of $\alpha$ being 1 and the rest being 0, and the lowest value if all $\alpha(i,j)$ have the same value.\footnote{Another regularization that can be used to fulfill the same purpose
of obtaining sharpening distributions 
 is the entropy $-\sum_{i,j} \alpha_{i,j}\log \alpha_{i,j}$ which can be viewed as the KL-divergence between the distribution and the uniform distribution. Empirically, we find that the two strategies perform almost the same.}
 The model  can be trained in an end-to-end fashion.
 \begin{comment}
  in two different ways: the end-to-end fashion and the pipelined fashion. 
For the former, we 
can directly train Eq.\ref{objective} in an end-to-end fashion, and use $\alpha(i,j)$ for interpretation. 
For the latter,
we can  first train a model based on vanilla transformers, BiLSTMs or CNNs to optimize $p(y|\bm{x})$,
 where the SIC layer and the interpretation layer are not used.
Next, we fix the input layer and the intermediate layers and only update parameters for the SIC layer and the interpretation layer based on Eq.\ref{objective}. 
\end{comment}
\section{Evaluation}
\label{sec:evaluation}
\begin{comment}
Though  substantial efforts have been invested to neural model interpretations, 
there has not been a generally accepted 
framework for evaluating different  interpretation models. 
One direct way for evaluation is to ask humans to annotate rationales within the input text, and compare  rationales extracted by interpretation models with human annotations, as in \newcite{deyoung-etal-2020-eraser}. 
The biggest issue with this methodology is that in many NLP tasks (e.g., fine-grained sentiment analysis), the difference between different labels is subtle (e.g.,  positive label v.s. very positive label or negative v.s. very negative label). It is thus hard for humans to annotate rationales that are both {\bf sufficient} and {\bf comprehensive}. 
Therefore, tasks in \newcite{deyoung-etal-2020-eraser}  are mostly limited to relatively easy NLP tasks, e.g., binary sentiment classification where 
for most data point, 
identifying constituent sentiment keywords suffices for predictions. 
\end{comment}
We would like an automatic evaluation method that is both flexible in evaluating the quality of model explanations under any task and dataset, and able to offer the ability of accurately reflecting the {\it faithfulness} of model explanations, i.e., the extracted rationales ought to semantically influence the model's prediction for the same instance,  as suggested by \citet{deyoung-etal-2020-eraser}.

Intuitively,  if  extracted rationales can {\bf faithfully} represent, in other words, be equivalent to, 
 their corresponding text inputs with respect to model predictions, the following should hold: 
(1)  a model trained on the original inputs 
   should perform comparably well when tested on the extracted rationales; 
   (2)  a model trained on the extracted rationales should  perform comparably well when tested on the original inputs;
   (3) a model trained on the  extracted rationales should  perform comparably well when tested on other extracted rationales.  
Higher performances on these three aspects indicate more faithful extracted rationales, and consequently better interpretation models. 
These strategies are inspired by \newcite{lei2016rationalizing}, who trained a rationale generator to extract text pieces as explanations for different sentiment aspects for the task of sentiment analysis, achieving high precision based on  sentence-level aspect annotations. 
   
   More formally, we use 
 {\it full} to refer to the situation of training or testing a model
on the original texts, 
 and  {\it span} to refer to the situation of training or testing a model
on the extracted rationales.  
By denoting the original full dataset by $D_\text{train}$, $D_\text{dev}$ and $D_\text{test}$, and the newly constructed rationale-based dataset by $D'_\text{train}$, $D'_\text{dev}$ and $D'_\text{test}$, the settings described above are as follows: 
\begin{itemize}
  \item {\bf FullTrain-SpanTest}: The model is trained on $D_\text{train}$ and tested on $D'_\text{test}$, with hyperparameters selected on $D_\text{dev}$.
  \item {\bf SpanTrain-FullTest}: The model is trained on $D'_\text{train}$ and tested on $D_\text{test}$, with hyperparameters selected on $D'_\text{dev}$.
  \item {\bf SpanTrain-SpanTest}: The model is trained on $D'_\text{train}$ and tested on $D'_\text{test}$, with hyperparameters selected on $D'_\text{dev}$.
\end{itemize}
To construct the new rationale-based dataset, we train a rationale-extraction model on $D_\text{train}$ to extract rationales from the original text in $D_\text{train}$, $D_\text{dev}$ and $D_\text{test}$, and then replace the original full input text with the corresponding extracted rationales, with the labels remaining unchanged.

It is also worthing that the proposed three evaluation metrics are far from perfect: the system can be gamed 
 if the extracted plan is just the same as the original span. The proposed evaluations thus need to be combined with other evaluations for complement.

\section{Experiments}
\subsection{Tasks and Datasets}
We conduct experiments on three NLP tasks: text classification on the SST-5 dataset \cite{socher2013recursive},
natural language inference on the SNLI dataset \cite{bowman2015large} 
and  machine translation on the IWSLT2014 En$\rightarrow$De dataset. 
Please refer to the appendix section for descriptions of the three datasets and training details.

For text classification and natural language inference, we use RoBERT \citep{yinhan2019roberta}
as the model backbone.  
For reference purposes, we also train a self-explaining model with $\bm{\alpha}$ set to be uniform, i.e., $\alpha(i,j)=1/M$, where $M$ is the total number of text spans. This model is denoted by AvgSelf-Explaining. 
For neural machine translation, we use the Transformer-base \citep{vaswani2017attention} model for evaluation.

\begin{table}[t]
  \centering
  \small
  \begin{tabular}{ll}
    \toprule
    {\bf Model} & {\bf Accuracy$\uparrow$}\\
    \hline\hline 
    \multicolumn{2}{c}{\textit{SST-5}}\\
    \midrule
    BCN+SuBiLSTM+CoVe \citep{brahma2018improved} & 56.2 \\
    BERT-base \citep{cheang2020language} & 54.9 \\
    BERT-large \citep{cheang2020language} & 56.2 \\
    SentiBERT \citep{yin2020sentibert}$^\flat$ & 56.9 \\
     SentiLARE \citep{ke-etal-2020-sentilare}$^{\flat\dag}$ & 58.6\\
    \cdashline{1-2}
    {RoBERTa-base+AvgSelf-Explaining} & 56.2\\
        {RoBERTa-base} \citep{yinhan2019roberta} & 56.4\\
    {RoBERTa-base+Self-Explaining} & {57.8 (+1.4)}\\
        \cdashline{1-2}
    {RoBERTa-large} \citep{yinhan2019roberta} & 57.9\\
    {RoBERTa-large+Self-Explaining} & {\bf 59.1 (+1.2)}\\
    \hline\hline 
    \multicolumn{2}{c}{\textit{SNLI}}\\
    \midrule
    BERT-base \citep{zhang2019explicit} & 89.2\\
    BERT-large \citep{zhang2019explicit} & 90.4\\
    SJRC \citep{zhang2019explicit}$^{\dag}$ & 91.3\\
    NILE \citep{kumar2020nile}$^\flat$ & 91.5\\
    MT-DNN \citep{liu-etal-2019-multi}$^{\dag}$ & 91.6\\
    SemBERT \citep{zhang2020semantics}$^{\dag}$ & 91.9 \\
     CA-MTL \citep{pilault2020conditionally}$^{\natural\dag}$ & 92.1 \\
    \cdashline{1-2}
    {RoBERTa-base+AvgSelf-Explaining} & 90.8\\
        {RoBERTa-base} \citep{yinhan2019roberta} & 90.7\\
    {RoBERTa-base+Self-Explaining} & {91.7 (+1.0)}\\   
    \cdashline{1-2}
    {RoBERTa-large} \citep{yinhan2019roberta} & 91.4\\
    {RoBERTa-large+Self-Explaining} & {\bf 92.3 (+0.9)}\\   
    \bottomrule
  \end{tabular}
  \caption{Performances of different models on the SST-5 and SNLI datasets. $^\flat$ denotes using RoBERTa-base as the model backbone, $^\natural$ denotes using RoBERTa-large as the model backbone, and  $^\dag$ denotes using external training resources.}
  \label{tab:main_results_1}
\end{table}

\begin{table}[t]
  \centering
  \small
  \begin{tabular}{ll}
    \toprule
    {\bf Model} & {\bf BLEU$\uparrow$}\\
    \hline\hline 
    \multicolumn{2}{c}{\textit{IWSLT 2014 En$\rightarrow$De}}\\
    \midrule
    Transformer-base \citep{vaswani2017attention} & 28.4\\
    Transformer-base+Self-Explaining & 28.9\\
    \bottomrule
  \end{tabular}
  \caption{Performances of different models on the IWSLT 2014 En$\rightarrow$De dataset.}
  \label{tab:main_results_2}
\end{table}

\subsection{Main Results}
Table \ref{tab:main_results_1} shows the results for the SST-5 and SNLI datasets, and Table \ref{tab:main_results_2} shows the results for IWSLT 2014 En$\rightarrow$De. We can see from the tables that the proposed Self-Explaining method  significantly boosts performances over strong RoBERTa baselines on SST-5 and SNLI. 
Using RoBerta-large,  we
 achieve a new SOTA performance of {\bf 59.1} on SST-5 and 
  a new SOTA performance of {\bf 92.3} 
 on SNLI. 
A surprising observation is that in spite of using the SIC layer and the interpretation layer, AvgSelf-Explaining still underperforms RoBERTa, which indicates that  the learned attention weights $\bm{\alpha}$ are important for model performances.

\subsection{Interpretability Evaluation}
We compare our proposed Self-Explaining model with the following three widely used interpretation models:
\begin{itemize}
    \item {\bf AvgGradient}: The method of averaging word saliency scores within a text span \citep{li2015visualizing,feng2018pathologies}.
    The saliency score of a word $w$ is computed as the derivative of the probability of the ouput label with respect to the word embedding: $s_y(w)=\|\nabla_{\bm{w}}p(y|\bm{x})\|_1$, where $p(y|\bm{x})$ is the predicted probability of the ground-truth label $y$, $\bm{w}$ is the corresponding word embedding of word $w$ and $\|\cdot\|_1$ is the L1 norm. We take the average of saliency scores of all words within a text span as the span-level saliency score. We select the span with the highest span-level saliency score as the rationale.
    \item {\bf AvgAttention}: The method of averaging attention scores within a text span \citep{vig2019analyzing,tenney2019bert,clark2019does}. The attention score for each word $w$ is the normalized attentive probability of the special token \texttt{[CLS]} with respect to
    word
     $w$ in the last intermediate layer. We take the average of attention scores of all words within a text span as the span-level attention score. The span with the highest 
     span-level 
     attention score is selected as the rationale.
    \item {\bf Rationale}: The rationale extraction model proposed by \citet{lei2016rationalizing}. One encoder is used to encode input texts into representative features, and one generator is used to extract text spans as rationales. These two models are jointly trained to minimize the expected cost function using the REINFORCE algorithm \citep{DBLP:journals/ml/Williams92}.

\end{itemize}
All models use RoBERTa \citep{yinhan2019roberta} as the backbone and are trained separately for comparison.

Results are shown in Table \ref{tab:evaluation}. For all the three setups FullTrain-SpanTest, SpanTrain-FullTest and SpanTrain-SpanTest, we can observe that the proposed Self-Explaining outperforms other interpretation methods by a large margin on both SST-5 and SNLI. On SST, Self-Explaining outperforms Rationale by +3.2, +5.0 and +6.7 respectively for the F-S, S-F and S-S setup. On SNLI, Self-Explaining outperforms Rationale by +3.5, +10.0 and 7.3 respectively for the F-S, S-F and S-S setup. 
Comparing AvgGradient, AvgAttention and Rationale, we can see that there is no method that shows consistent  superiority to the other two. For example, Rationale outperforms AvgGradient and AvgAttention under the F-S and S-F setup, but underperforms AvgGradient under the S-S setup. Instead, Self-Explaining consistently achieves significantly better results compared to all baselines  under all setups.
These results demonstrate the better interpretability of the proposed Self-Explaining method.

\begin{comment}
\subsection{Evaluation on ERASER}
We also evaluate interpretability performances of different models on the ERASER dataset \citep{deyoung-etal-2020-eraser}. 
Evaluations are conducted on the 
Movie Review dataset and SNLI dataset in ERASER.  
Since the dataset is labeled in a way that an input can contain multiple rationales, we treat the threshold for $\alpha$ as the hyper-parameter to be tuned on the dev sets of ERASER.
Spans with $\alpha$ higher than the threshold will be selected. To deal with the issue of overlapping spans, we adopt the following strategy: we select spans in a top-down fashion, and the current span will be ignored if it overlaps with previously selected spans. 
We compared the extracted spans with the golden labeled spans and report  IOU F1 and token F1 scores. 
Results are shown in Table \ref{tab:eraser}. Performances on both the Movie Review and the SNLI datasets exhibit significant improvements of Self-Explaining over other baselines, demonstrating the stronger ability of extracting rationales of the proposed Self-Explaining method.
\end{comment}

\begin{table}[t]
  \centering
  \small
  \begin{tabular}{lll}
    \toprule
    {\bf Model} & {\bf IOU F1} & {\bf Token F1}\\
    \hline\hline 
    \multicolumn{3}{c}{\textit{Movie Review}}\\
    \midrule
    AvgGradient & 0.075 & 0.175 \\ 
    AvgAttention & 0.067 & 0.142  \\
    Rationale \citep{lei2016rationalizing} & 0.132  & 0.281 \\
    Self-Explaining & {\bf 0.152} & {\bf 0.314}  \\
    \hline\hline 
    \multicolumn{3}{c}{\textit{SNLI}}\\
    \midrule
    AvgGradient & 0.251& 0.349  \\ 
    AvgAttention & 0.301 &  0.435 \\
    Rationale \citep{lei2016rationalizing} & 0.381 & 0.459 \\
    Self-Explaining & {\bf 0.454} & {\bf 0.571}  \\
    \bottomrule
  \end{tabular}
  \caption{Performances of different models on Movie Review and SNLI of the ERASER benchmark.}
  \label{tab:eraser}
\end{table}

\begin{table}[t]
  \centering
  \small
  \begin{tabular}{llll}
    \toprule
    {\bf Model} & {\bf F-S} & {\bf S-F} & {\bf S-S}\\
    \hline\hline 
    \multicolumn{4}{c}{\textit{SST-5}}\\
    \midrule
    AvgGradient & 34.1 & 45.6 & 36.9 \\ 
    AvgAttention & 32.5 & 40.6 & 35.8 \\
    Rationale \citep{lei2016rationalizing} & 35.2 & 49.9 & 36.2\\
    Self-Explaining & {\bf 38.4} & {\bf 54.9} & {\bf 42.9} \\
    \hline\hline 
    \multicolumn{4}{c}{\textit{SNLI}}\\
    \midrule
    AvgGradient & 70.7 & 74.5 & 73.1 \\ 
    AvgAttention & 64.5 & 72.5 & 70.4 \\
    Rationale \citep{lei2016rationalizing} & 71.0 & 78.5 & 71.9\\
    Self-Explaining & {\bf 74.5} & {\bf 88.5} & {\bf 79.2} \\
    \bottomrule
  \end{tabular}
  \caption{Performances of different models on the three evaluation methods defined in Section \ref{sec:evaluation}. ``F'' refers to {\it Full} and ``S'' refers to {\it Span}. Accuracy is reported in each cell.}
  \label{tab:evaluation}
\end{table}

\begin{table*}[t]
    \centering
    \small
    \begin{tabular}{llp{12cm}}
    \toprule
       {\bf Label}  & {\bf Model} & {\bf Text}   \\
     \midrule
    \multirow{3}{*}{{\it Very Negative}} & (1) &this {\bf overproduced and generally disappointing} effort isn't likely to rouse the rush hour crowd  \\
    & (2) &this overproduced and generally {\bf disappointing} effort isn't likely to rouse the rush hour crowd\\
    & (3)& this {\bf overproduced and generally disappointing} effort isn't likely to rouse the rush hour crowd\\
    \cline{1-3}
    \multirow{3}{*}{{\it Negative}} &  (1)& However, {\bf it lacks grandeur and that epic quality often associated with Stevenson's tale} as well as with earlier Disney efforts  \\
    & (2) &However, it {\bf lacks grandeur} and that epic quality often associated with Stevenson's tale as well as with earlier Disney efforts \\
    &(3) & However, it {\bf lacks grandeur} and that epic quality often associated with Stevenson's tale] as well as with earlier Disney efforts\\
    \cline{1-3}
    \multirow{3}{*}{{\it Negative}} & (1)&  {\bf Though everything might be literate and smart}, it never took off and always seemed static \\
    & (2) &Though everything might be literate and {\bf smart}, it never took off and always seemed static\\
    & (3) &Though everything might be {\bf literate and smart}, it never took off and always seemed static \\
    \cline{1-3}
    \multirow{3}{*}{{\it Very Positive}} &(1) & {\bf One of the greatest family-oriented, fantasy-adventure movies ever } \\
    & (2) &One of the {\bf greatest} family-oriented, fantasy-adventure movies ever \\
    & (3) &One of the {\bf greatest} family-oriented, fantasy-adventure movies ever \\
    \bottomrule
    \end{tabular}
    \caption{Examples of correctly classified texts and the corresponding {\bf extracted text spans} by different models. (1): Self-Explaining; (2): AvgGradient; (3): AvgAttention. }
    \label{tab:example}
\end{table*}

\begin{figure}
  \centering
  \includegraphics[scale=0.4]{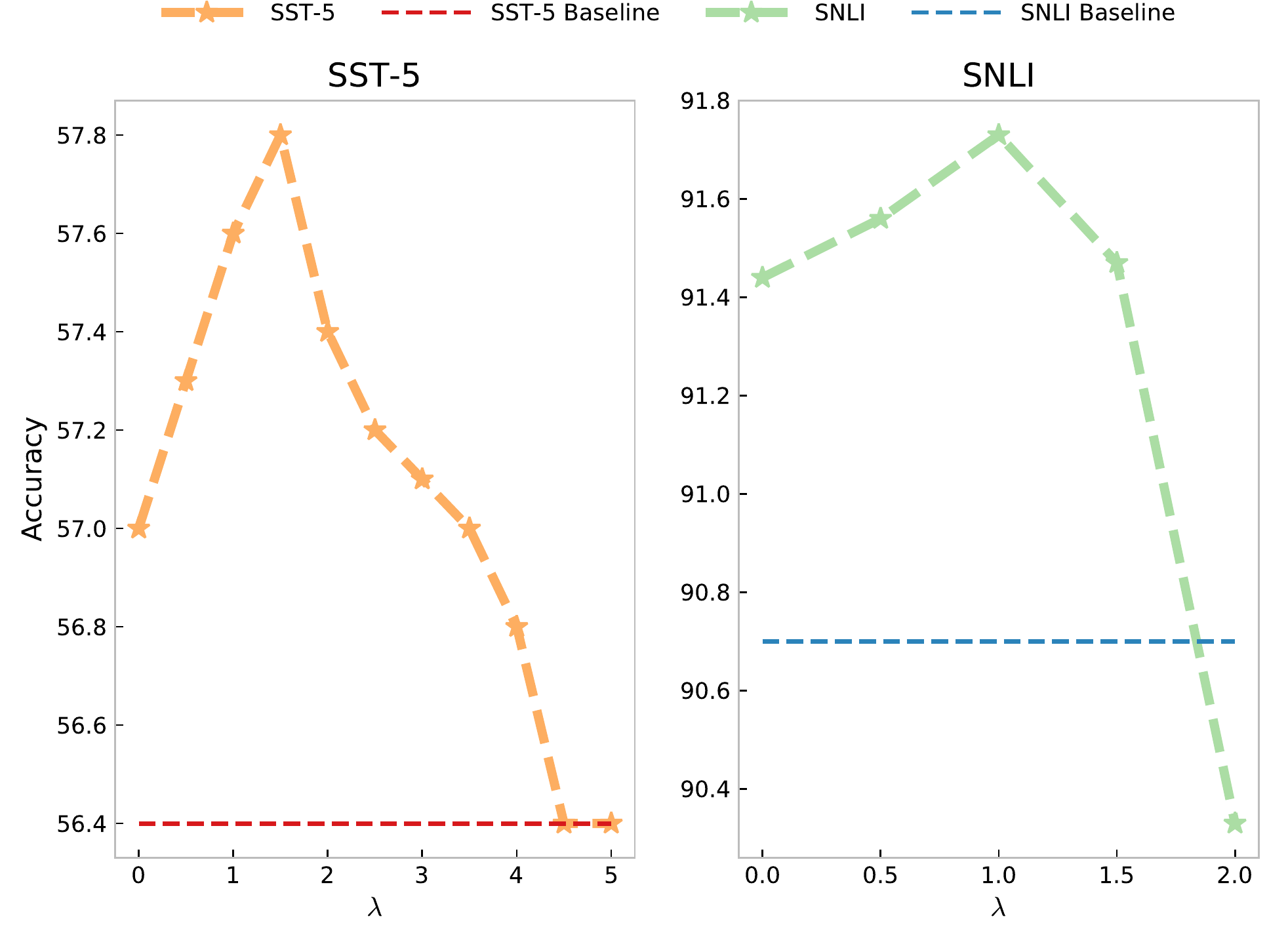}
  \caption{Performances of the proposed model with respect to different values of $\lambda$. Accuracy is reported for comparison.}
  \label{fig:lambda}
\end{figure}

\subsection{The Effect of $\lambda$}
We would like to explore the effect of $\lambda$ in the additional regularization term in Eq.\ref{objective}. A larger $\lambda$ has a stronger effect on sharpening the distribution of $\bm{\alpha}$, leading to the preference to a small fraction of text spans. Intuitively, a reasonable value of $\lambda$ is crucial for model performances: too small $\lambda$  means the selected text spans are not confident enough to support model predictions, while too large $\lambda$ means the model only attends to a single text span for its prediction. The former may cause  over-numbered selected text spans, which incurs more noise for model predictions. The latter may force the model the attend to a single span that is not important for predictions. Therefore, a sensible $\lambda$ should be neither too small nor too large, balancing the number of attended text spans.

Figure \ref{fig:lambda} shows the results. As we can see from the figure, $\lambda=1.5$ leads to the best result on SST-5 and $\lambda=1.0$ gives the best result on SNLI, significantly outperforming models with small $\lambda$ and large $\lambda$, as well as the baseline. As the value of $\lambda$ keeps increasing, the performance drastically drops. When $\lambda=5.0$ for SST-5 and $\lambda=2.0$ for SNLI, the accuracies are respectively 56.4 and 90.3, even underperforming baselines without self-explaining structures. 
To our surprise, though the model performance with $\lambda=0$ is worse than the best model with $\lambda=1.5$ and $\lambda=1.0$, it still achieves better results compared to baselines (57.0 vs. 56.4 on SST-5 and 91.4 vs. 90.7 on SNLI). This observation verifies that the proposed method can actually improve NLP models in terms of both performance and interpretability.

\begin{table*}[t]
  \centering
  \small
  \begin{tabular}{llp{11.5cm}}
  \toprule
  {\bf True} & {\bf Predicted} & {\bf Text} \\
   \midrule
   {\it Negative} & {\it Positive} & {{\bf the story is naturally poignant}, but first - time screen-writer paul pender overloads it with sugary bits ofbusiness}\\
   {\it Negative} & {\it Neutral} & {{\bf a well acted and well intentioned} snoozer}\\
   {\it Negative} & {\it Positive} & {george, {\bf hire a real director} and good writers forthe next installment, please}\\
   {\it Positive} & {\it Negative} & {It 's {\bf like a poem}}\\
   {\it Positive} & {\it Very Negative} & {There {\bf isn't a weak or careless performance} amongst them}\\
   \bottomrule
  \end{tabular}
  \caption{Examples of mis-classified texts and the corresponding {\bf extracted text spans}.}
  \label{tab:error}
\end{table*}
\begin{table*}[!ht]
  \centering
  \small
  \begin{tabular}{p{1.5cm}p{13.5cm}}
  \toprule
  {\bf Label}  & {\bf Text} \\
   \midrule
   \makecell[c]{{\it Positive}\\{{$\hookrightarrow$}\it Negative}} & \multicolumn{1}{m{13.5cm}}{In the film, Lumumba, we see the faces behind the monumental shift in the Congo's history after it is  reclaimed from the Belgians, and we see the motives behind those men into whose hands the raped and         starving country fell. Lumumba is not a movie for the hyper masses; it demands the  attention of its viewers with raw, truthful acting and intricate, packed dialogue. Little of the main   plot is shown through action, it relies almost solely on words, but there is a recurring strand that is only action, and it is the stroke of genius that makes the film an enlightening and powerful panorama  of the tense political struggle that the Congo's independence gave birth to.This film is   real. It is raw inits depiction of those in power, and those on the streets. It is eye-opening in its  content. And it is moving in the passions and emotions of its superbly portrayed characters. {\bf Whether you are a history fan, a film buff, or simply like good stories, Lumumba is a must-see {\color{red}[Whether you are a history fan, a cinephile or just love good stories, Lumumba is a must-have]}}}.\\
  \midrule
  \makecell[c]{{\it Negative}\\{{$\hookrightarrow$}\it Positive}} & \multicolumn{1}{m{13.5cm}}{{\bf I'm a huge Zack Allan fan and was disappointed that he only got one scene in the movie {\color{red}[I'm a huge fan of Zack Allan and I was disappointed that he only has one scene in the film]}}. This was also my favourite scene where he confiscates a character's weapons and directs her to Down Below. Unfortunately unlike Thirdspace \& River of Souls, most of the action took place off station. I didn't care much for Garibaldi after the first three seasons and think Sheridan is okay but no Sinclair. I like Lochley but she only had limited screen time. If you like Crusade or space battles you should enjoy it.}\\
  \bottomrule
  \end{tabular}
  \caption{Examples of generated adversarial text spans on the SST-5 dataset. The extracted and back-translated text spans are in {\bf bold}. The text in  parentheses {\color{red}[$\cdot$]} is the corresponding paraphrase to flip the prediction.}
  \label{tab:adversarial_example}
\end{table*}

\section{Analysis}
\subsection{Examples}
Table \ref{tab:example} lists several examples to illustrate how different  methods extract text spans to interpret model predictions. Extracted spans are in bold. 

As can be seen from the table, all methods including Self-Explaining are able to extract corresponding text spans as evidential explanation for the model prediction. For example, both AvgAttention and the proposed Self-Explaining are able to extract ``overproduced and generally disappointing'' as rationale for predicting the label {\it Very Negative}, and AvgGradient selects the key term ``disappointing'', which also makes sense in this case.
However, compared to AvgGradient and AvgAttention, Self-Explaining can select more global and comprehensive text spans. For example, Self-Explaining extracts ``it lacks grandeur and that epic quality often associated with Stevenson’s tale'' for predicting the {\it Negative} label, while AvgAttention and AvgGradient only extract ``lacks grandeur'', which may not be comprehensive for making the decision. Self-Explaining is able to extract ``Though everything might be literate and smart'' as rationale for predicting {\it Negative}, while AvgAttention and AvgGradient only extract local text spans ``smart'' and ``literate and smart'',  respectively. Although they all  make correct predictions, the rationales provided by  AvgAttention and AvgGradient can not explain their behaviors of making the decision. By contrast, Self-Explaining successfully captures the conjunction ``though'', a word that  reverses the sentiment from positive to negative.

\subsection{Error Analysis}
By examining extracted text spans from erroneously classified examples, 
the model provides a direct way for performing error analysis.
The biggest advantage of the proposed model over previous interpretability methods is that, instead of  focusing merely on word-level features as in \citet{li2015visualizing,li2016understanding},
the proposed model operates at arbitrary levels of text units, providing more direct and accurate views of why models make mistakes. 
 By examining erroneously classified examples shown in Table \ref{tab:error}, 
we can clearly identify 
a few patterns that make neural models fail: 
(1) the model 
emphasizes the  part of sentences that should not be attended to in the 
contrast conjunction, e.g., ``
[the story is naturally poignant], but first - time screenwriter paul pender overloads it with sugary bits of business'';
(2) the model cannot recognize  a word used in a context that changes its sentiment, e.g., ``
[a well acted and well intention] ed snoozer'' and ``There [isn't a weak or careless performance] amongst them'';
(3) the model cannot recognize irony: 
``george, [hire a real director] and good writers for the next installment, please''; (4) the model cannot recognize analogy: ``It's [like a poem]'', etc. 

\begin{table}[t]
  \centering
  \small
  \begin{tabular}{lcc}
    \toprule
    {\bf Model} & {\it IMBD} & {\it Yahoo!Answers} \\
    \midrule
    Original & 84.86&92.00 \\
    \cdashline{1-3}[0.8pt/2pt]
    Random & 37.79&74.50 \\
    Gradient &14.57&73.80 \\
    TiWo & 3.57&62.50 \\
    WS & 3.93&62.50 \\
    PWWS & 2.00&53.00 \\
    Self-Explaining+Paraphrase &{\bf 0.86} &{\bf 43.14} \\
    \bottomrule
  \end{tabular}
  \caption{Classification accuracy of each model on the original datasets (the first row) and the perturbed datasets using different adversarial methods. A lower classification accuracy corresponds to a more effective attacking method.}
  \label{tab:adversarial}
\end{table}

\subsection{Span-based Adversarial Example Generation}
There has been a growing interest in generating adversarial examples to attack a neural network model in NLP \citep{alzantot2018generating,ren-etal-2019-generating,zhang2020adversarial}. 
The current  protocol for adversarial example generation is based on word substitution \citep{liang2017deep,ebrahimi2017hotflip,samanta2017towards,ren-etal-2019-generating}, where an important word is  selected by saliency  and then 
replaced by its synonym if the replacement can flip the prediction. 
The shortcoming for current approaches is that it can only operate at the word level and perform word-level substitutions. This is rooted in the fact that 
 saliency scores can only be reliably computed at the word level.

The proposed span-based model naturally addresses this issue: 
we  first  identify the most salient span in the input based on $\alpha$, replace it with its paraphrase if the replacement flips the label.
To the best of our knowledge, this is the first feasible approach that operates at higher levels beyond words for adversarial example generation in NLP. 
Paraphrases are generated by 
back-translation \citep{sennrich2016back-translation,edunov2018understanding}.\footnote{We use the 
off-the-shelf google translator to implement En$\rightarrow$De$\rightarrow$En back-translation.} 

Following \citet{ren-etal-2019-generating}, 
we use two datasets for test, IMDB \citep{maas2011learning} and Yahoo! Answers.
The details of the two datasets are described in Appendix. 
For fair comparisons, we follow \citet{ren-etal-2019-generating} to use the Bi-LSTMs as the model backbone and compare our proposed Self-Explaining+Paraphrase model with the following attacking methods: (1) Random; (2) {Gradient}; (3) {Traversing in Word Order (TiWO)}; (4) {Word Saliency (WS)}; (5) Probability Weighted Word Saliency (PWWS). We refer readers to  \citet{ren-etal-2019-generating} for more details of these methods.

Table \ref{tab:adversarial} shows the classification accuracies of different methods on the original datasets and the adversarial samples generated by these attacking methods. 
Results show that our proposed Self-Explaining+Paraphrase method reduces the classification accuracy to the most extent. Compared to the original performance, Self-Explaining+Paraphrase reduces the classification accuracies on IMDB and Yahoo! Answers by 84\% and 48.86\%, respectively, indicating the effectiveness of the proposed Self-Explaining+Paraphrase method for generating adversarial samples.

Table \ref{tab:adversarial_example} gives two examples of flipping model predictions by using the Self-Explaining+Paraphrase method to generate adversarial text spans. Through the examples, we can see that the selected text span is crucial for the model prediction and a slight perturbation injected to this span would flip the prediction.

\section{Conclusion}
In this work, we present a light but effective self-explainable structure to improve both performance and interpretability in the context of NLP.  The idea of the proposed method is to introduce an interpretation layer, aggregating information for each text span, which is then assigned a specific weight representing its contribution to interpreting  the model prediction. %In this way, the model can be trained to simultaneously perform downstream tasks and explain its behaviors at a higher level such as phrase and sentence. 
Extensive experiments show the effectiveness of the proposed method in terms of improving performances for the task of sentiment classification and natural language inference, as well as endowing the model with the ability of self-explaining, avoiding heavy recourse to additional external interpretation models such as probing models and surrogate models. 
%In future work, we would like to discover how the proposed method is connected to existing interpretation methods and apply it to more NLP tasks.

\bibliography{emnlp2020}

\begin{thebibliography}{76}
\expandafter\ifx\csname natexlab\endcsname\relax\def\natexlab#1{#1}\fi

\bibitem[{Adler et~al.(2018)Adler, Falk, Friedler, Nix, Rybeck, Scheidegger,
  Smith, and Venkatasubramanian}]{adler2018auditing}
Philip Adler, Casey Falk, Sorelle~A Friedler, Tionney Nix, Gabriel Rybeck,
  Carlos Scheidegger, Brandon Smith, and Suresh Venkatasubramanian. 2018.
\newblock Auditing black-box models for indirect influence.
\newblock \emph{Knowledge and Information Systems}, 54(1):95--122.

\bibitem[{Alzantot et~al.(2018)Alzantot, Sharma, Elgohary, Ho, Srivastava, and
  Chang}]{alzantot2018generating}
Moustafa Alzantot, Yash Sharma, Ahmed Elgohary, Bo-Jhang Ho, Mani Srivastava,
  and Kai-Wei Chang. 2018.
\newblock Generating natural language adversarial examples.
\newblock \emph{arXiv preprint arXiv:1804.07998}.

\bibitem[{Arras et~al.(2016)Arras, Horn, Montavon, M{\"u}ller, and
  Samek}]{arras2016explaining}
Leila Arras, Franziska Horn, Gr{\'e}goire Montavon, Klaus-Robert M{\"u}ller,
  and Wojciech Samek. 2016.
\newblock Explaining predictions of non-linear classifiers in nlp.
\newblock \emph{arXiv preprint arXiv:1606.07298}.

\bibitem[{Arrieta et~al.(2020)Arrieta, D{\'\i}az-Rodr{\'\i}guez, Del~Ser,
  Bennetot, Tabik, Barbado, Garc{\'\i}a, Gil-L{\'o}pez, Molina, Benjamins
  et~al.}]{arrieta2020explainable}
Alejandro~Barredo Arrieta, Natalia D{\'\i}az-Rodr{\'\i}guez, Javier Del~Ser,
  Adrien Bennetot, Siham Tabik, Alberto Barbado, Salvador Garc{\'\i}a, Sergio
  Gil-L{\'o}pez, Daniel Molina, Richard Benjamins, et~al. 2020.
\newblock Explainable artificial intelligence (xai): Concepts, taxonomies,
  opportunities and challenges toward responsible ai.
\newblock \emph{Information Fusion}, 58:82--115.

\bibitem[{Bach et~al.(2015{\natexlab{a}})Bach, Binder, Montavon, Klauschen,
  M{\"u}ller, and Samek}]{bach2015lrp}
Sebastian Bach, Alexander Binder, Gr{\'e}goire Montavon, Frederick Klauschen,
  Klaus-Robert M{\"u}ller, and Wojciech Samek. 2015{\natexlab{a}}.
\newblock On pixel-wise explanations for non-linear classifier decisions by
  layer-wise relevance propagation.
\newblock \emph{PloS one}, 10(7).

\bibitem[{Bach et~al.(2015{\natexlab{b}})Bach, Binder, Montavon, Klauschen,
  Müller, and Samek}]{10.1371/journal.pone.0130140}
Sebastian Bach, Alexander Binder, Grégoire Montavon, Frederick Klauschen,
  Klaus-Robert Müller, and Wojciech Samek. 2015{\natexlab{b}}.
\newblock On pixel-wise explanations for non-linear classifier decisions by
  layer-wise relevance propagation.
\newblock \emph{PLOS ONE}, 10:1--46.

\bibitem[{Bahdanau et~al.(2014)Bahdanau, Cho, and Bengio}]{bahdanau2014neural}
Dzmitry Bahdanau, Kyunghyun Cho, and Yoshua Bengio. 2014.
\newblock Neural machine translation by jointly learning to align and
  translate.

\bibitem[{Bowman et~al.(2015)Bowman, Angeli, Potts, and
  Manning}]{bowman2015large}
Samuel~R Bowman, Gabor Angeli, Christopher Potts, and Christopher~D Manning.
  2015.
\newblock A large annotated corpus for learning natural language inference.
\newblock \emph{arXiv preprint arXiv:1508.05326}.

\bibitem[{Brahma(2018)}]{brahma2018improved}
Siddhartha Brahma. 2018.
\newblock Improved sentence modeling using suffix bidirectional lstm.
\newblock \emph{arXiv preprint arXiv:1805.07340}.

\bibitem[{Chai et~al.(2020)Chai, Wu, Han, Wu, and Li}]{chai2020description}
Duo Chai, Wei Wu, Qinghong Han, Fei Wu, and Jiwei Li. 2020.
\newblock Description based text classification with reinforcement learning.

\bibitem[{Chang et~al.(2019)Chang, Zhang, Yu, and Jaakkola}]{chang2019game}
Shiyu Chang, Yang Zhang, Mo~Yu, and Tommi Jaakkola. 2019.
\newblock A game theoretic approach to class-wise selective rationalization.
\newblock In \emph{Advances in Neural Information Processing Systems}, pages
  10055--10065.

\bibitem[{Cheang et~al.(2020)Cheang, Wei, Kogan, Qiu, and
  Ahmed}]{cheang2020language}
Brian Cheang, Bailey Wei, David Kogan, Howey Qiu, and Masud Ahmed. 2020.
\newblock Language representation models for fine-grained sentiment
  classification.
\newblock \emph{arXiv preprint arXiv:2005.13619}.

\bibitem[{Chen et~al.(2019)Chen, Khashabi, Yin, Callison-Burch, and
  Roth}]{chen-etal-2019-seeing}
Sihao Chen, Daniel Khashabi, Wenpeng Yin, Chris Callison-Burch, and Dan Roth.
  2019.
\newblock Seeing things from a different angle:discovering diverse perspectives
  about claims.
\newblock In \emph{Proceedings of the 2019 Conference of the North {A}merican
  Chapter of the Association for Computational Linguistics: Human Language
  Technologies, Volume 1 (Long and Short Papers)}, Minneapolis, Minnesota.
  Association for Computational Linguistics.

\bibitem[{Clark et~al.(2019)Clark, Khandelwal, Levy, and
  Manning}]{clark2019does}
Kevin Clark, Urvashi Khandelwal, Omer Levy, and Christopher~D. Manning. 2019.
\newblock What does bert look at? an analysis of bert's attention.

\bibitem[{Datta et~al.(2016)Datta, Sen, and Zick}]{datta2016algorithmic}
Anupam Datta, Shayak Sen, and Yair Zick. 2016.
\newblock Algorithmic transparency via quantitative input influence: Theory and
  experiments with learning systems.
\newblock In \emph{2016 IEEE symposium on security and privacy (SP)}, pages
  598--617. IEEE.

\bibitem[{Denil et~al.(2015)Denil, Demiraj, and
  de~Freitas}]{denil2015extraction}
Misha Denil, Alban Demiraj, and Nando de~Freitas. 2015.
\newblock Extraction of salient sentences from labelled documents.

\bibitem[{DeYoung et~al.(2020)DeYoung, Jain, Rajani, Lehman, Xiong, Socher, and
  Wallace}]{deyoung-etal-2020-eraser}
Jay DeYoung, Sarthak Jain, Nazneen~Fatema Rajani, Eric Lehman, Caiming Xiong,
  Richard Socher, and Byron~C. Wallace. 2020.
\newblock {ERASER}: {A} benchmark to evaluate rationalized {NLP} models.
\newblock In \emph{Proceedings of the 58th Annual Meeting of the Association
  for Computational Linguistics}, Online. Association for Computational
  Linguistics.

\bibitem[{Ebrahimi et~al.(2017)Ebrahimi, Rao, Lowd, and
  Dou}]{ebrahimi2017hotflip}
Javid Ebrahimi, Anyi Rao, Daniel Lowd, and Dejing Dou. 2017.
\newblock Hotflip: White-box adversarial examples for text classification.
\newblock \emph{arXiv preprint arXiv:1712.06751}.

\bibitem[{Edunov et~al.(2018)Edunov, Ott, Auli, and
  Grangier}]{edunov2018understanding}
Sergey Edunov, Myle Ott, Michael Auli, and David Grangier. 2018.
\newblock Understanding back-translation at scale.
\newblock \emph{arXiv preprint arXiv:1808.09381}.

\bibitem[{Feng et~al.(2018)Feng, Wallace, Grissom~II, Iyyer, Rodriguez, and
  Boyd-Graber}]{feng2018pathologies}
Shi Feng, Eric Wallace, Alvin Grissom~II, Mohit Iyyer, Pedro Rodriguez, and
  Jordan Boyd-Graber. 2018.
\newblock Pathologies of neural models make interpretations difficult.
\newblock \emph{arXiv preprint arXiv:1804.07781}.

\bibitem[{Ghaeini et~al.(2018)Ghaeini, Fern, and
  Tadepalli}]{ghaeini2018interpreting}
Reza Ghaeini, Xiaoli~Z Fern, and Prasad Tadepalli. 2018.
\newblock Interpreting recurrent and attention-based neural models: a case
  study on natural language inference.
\newblock \emph{arXiv preprint arXiv:1808.03894}.

\bibitem[{Greff et~al.(2016)Greff, Srivastava, Koutn{\'\i}k, Steunebrink, and
  Schmidhuber}]{greff2016lstm}
Klaus Greff, Rupesh~K Srivastava, Jan Koutn{\'\i}k, Bas~R Steunebrink, and
  J{\"u}rgen Schmidhuber. 2016.
\newblock Lstm: A search space odyssey.
\newblock \emph{IEEE transactions on neural networks and learning systems},
  28(10):2222--2232.

\bibitem[{Htut et~al.(2019)Htut, Phang, Bordia, and Bowman}]{htut2019attention}
Phu~Mon Htut, Jason Phang, Shikha Bordia, and Samuel~R. Bowman. 2019.
\newblock Do attention heads in bert track syntactic dependencies?

\bibitem[{Jain and Wallace(2019)}]{jain2019attention}
Sarthak Jain and Byron~C. Wallace. 2019.
\newblock Attention is not explanation.

\bibitem[{Jain et~al.(2020)Jain, Wiegreffe, Pinter, and
  Wallace}]{jain-etal-2020-learning}
Sarthak Jain, Sarah Wiegreffe, Yuval Pinter, and Byron~C. Wallace. 2020.
\newblock {L}earning to faithfully rationalize by construction.
\newblock In \emph{Proceedings of the 58th Annual Meeting of the Association
  for Computational Linguistics}, Online. Association for Computational
  Linguistics.

\bibitem[{Joshi et~al.(2020)Joshi, Chen, Liu, Weld, Zettlemoyer, and
  Levy}]{joshi2020spanbert}
Mandar Joshi, Danqi Chen, Yinhan Liu, Daniel~S Weld, Luke Zettlemoyer, and Omer
  Levy. 2020.
\newblock Spanbert: Improving pre-training by representing and predicting
  spans.
\newblock \emph{Transactions of the Association for Computational Linguistics},
  8:64--77.

\bibitem[{Karpathy et~al.(2015)Karpathy, Johnson, and
  Fei-Fei}]{karpathy2015visualizing}
Andrej Karpathy, Justin Johnson, and Li~Fei-Fei. 2015.
\newblock Visualizing and understanding recurrent networks.
\newblock \emph{arXiv preprint arXiv:1506.02078}.

\bibitem[{Ke et~al.(2020)Ke, Ji, Liu, Zhu, and Huang}]{ke-etal-2020-sentilare}
Pei Ke, Haozhe Ji, Siyang Liu, Xiaoyan Zhu, and Minlie Huang. 2020.
\newblock {S}enti{LARE}: Sentiment-aware language representation learning with
  linguistic knowledge.
\newblock In \emph{Proceedings of the 2020 Conference on Empirical Methods in
  Natural Language Processing (EMNLP)}, pages 6975--6988, Online. Association
  for Computational Linguistics.

\bibitem[{Kindermans et~al.(2017)Kindermans, Schütt, Alber, Müller, Erhan,
  Kim, and Dähne}]{kindermans2017learning}
Pieter-Jan Kindermans, Kristof~T. Schütt, Maximilian Alber, Klaus-Robert
  Müller, Dumitru Erhan, Been Kim, and Sven Dähne. 2017.
\newblock Learning how to explain neural networks: Patternnet and
  patternattribution.

\bibitem[{Kingma and Ba(2014)}]{kingma2014adam}
Diederik~P Kingma and Jimmy Ba. 2014.
\newblock Adam: A method for stochastic optimization.
\newblock \emph{arXiv preprint arXiv:1412.6980}.

\bibitem[{Koh and Liang(2017)}]{koh2017understanding}
Pang~Wei Koh and Percy Liang. 2017.
\newblock Understanding black-box predictions via influence functions.
\newblock In \emph{Proceedings of the 34th International Conference on Machine
  Learning-Volume 70}, pages 1885--1894. JMLR. org.

\bibitem[{Kumar and Talukdar(2020)}]{kumar2020nile}
Sawan Kumar and Partha Talukdar. 2020.
\newblock Nile: Natural language inference with faithful natural language
  explanations.
\newblock \emph{arXiv preprint arXiv:2005.12116}.

\bibitem[{Lehman et~al.(2019)Lehman, DeYoung, Barzilay, and
  Wallace}]{lehman-etal-2019-inferring}
Eric Lehman, Jay DeYoung, Regina Barzilay, and Byron~C. Wallace. 2019.
\newblock Inferring which medical treatments work from reports of clinical
  trials.
\newblock In \emph{Proceedings of the 2019 Conference of the North {A}merican
  Chapter of the Association for Computational Linguistics: Human Language
  Technologies, Volume 1 (Long and Short Papers)}, Minneapolis, Minnesota.
  Association for Computational Linguistics.

\bibitem[{Lei et~al.(2016)Lei, Barzilay, and Jaakkola}]{lei2016rationalizing}
Tao Lei, Regina Barzilay, and Tommi Jaakkola. 2016.
\newblock Rationalizing neural predictions.
\newblock \emph{arXiv preprint arXiv:1606.04155}.

\bibitem[{Li et~al.(2015)Li, Chen, Hovy, and Jurafsky}]{li2015visualizing}
Jiwei Li, Xinlei Chen, Eduard Hovy, and Dan Jurafsky. 2015.
\newblock Visualizing and understanding neural models in nlp.
\newblock \emph{arXiv preprint arXiv:1506.01066}.

\bibitem[{Li et~al.(2016)Li, Monroe, and Jurafsky}]{li2016understanding}
Jiwei Li, Will Monroe, and Dan Jurafsky. 2016.
\newblock Understanding neural networks through representation erasure.
\newblock \emph{arXiv preprint arXiv:1612.08220}.

\bibitem[{Li et~al.(2019)Li, Feng, Meng, Han, Wu, and Li}]{li2019unified}
Xiaoya Li, Jingrong Feng, Yuxian Meng, Qinghong Han, Fei Wu, and Jiwei Li.
  2019.
\newblock A unified mrc framework for named entity recognition.
\newblock \emph{arXiv preprint arXiv:1910.11476}.

\bibitem[{Liang et~al.(2017)Liang, Li, Su, Bian, Li, and Shi}]{liang2017deep}
Bin Liang, Hongcheng Li, Miaoqiang Su, Pan Bian, Xirong Li, and Wenchang Shi.
  2017.
\newblock Deep text classification can be fooled.
\newblock \emph{arXiv preprint arXiv:1704.08006}.

\bibitem[{Liu et~al.(2019{\natexlab{a}})Liu, He, Chen, and
  Gao}]{liu-etal-2019-multi}
Xiaodong Liu, Pengcheng He, Weizhu Chen, and Jianfeng Gao. 2019{\natexlab{a}}.
\newblock Multi-task deep neural networks for natural language understanding.
\newblock In \emph{Proceedings of the 57th Annual Meeting of the Association
  for Computational Linguistics}, pages 4487--4496, Florence, Italy.
  Association for Computational Linguistics.

\bibitem[{Liu et~al.(2019{\natexlab{b}})Liu, Ott, Goyal, Du, Joshi, Chen, Levy,
  Lewis, Zettlemoyer, and Stoyanov}]{yinhan2019roberta}
Yinhan Liu, Myle Ott, Naman Goyal, Jingfei Du, Mandar Joshi, Danqi Chen, Omer
  Levy, Mike Lewis, Luke Zettlemoyer, and Veselin Stoyanov. 2019{\natexlab{b}}.
\newblock Roberta: A robustly optimized bert pretraining approach.
\newblock \emph{arXiv preprint arXiv:1907.11692}.

\bibitem[{Maas et~al.(2011)Maas, Daly, Pham, Huang, Ng, and
  Potts}]{maas2011learning}
Andrew Maas, Raymond~E Daly, Peter~T Pham, Dan Huang, Andrew~Y Ng, and
  Christopher Potts. 2011.
\newblock Learning word vectors for sentiment analysis.
\newblock In \emph{Proceedings of the 49th annual meeting of the association
  for computational linguistics: Human language technologies}, pages 142--150.

\bibitem[{Melis and Jaakkola(2018)}]{melis2018towards}
David~Alvarez Melis and Tommi Jaakkola. 2018.
\newblock Towards robust interpretability with self-explaining neural networks.
\newblock In \emph{Advances in Neural Information Processing Systems}, pages
  7775--7784.

\bibitem[{Meng et~al.(2020)Meng, Fan, Sun, Hovy, Wu, and Li}]{meng2020pair}
Yuxian Meng, Chun Fan, Zijun Sun, Eduard Hovy, Fei Wu, and Jiwei Li. 2020.
\newblock Pair the dots: Jointly examining training history and test stimuli
  for model interpretability.
\newblock \emph{arXiv preprint arXiv:2010.06943}.

\bibitem[{Montavon et~al.(2017)Montavon, Lapuschkin, Binder, Samek, and
  M{\"u}ller}]{montavon2017explaining}
Gr{\'e}goire Montavon, Sebastian Lapuschkin, Alexander Binder, Wojciech Samek,
  and Klaus-Robert M{\"u}ller. 2017.
\newblock Explaining nonlinear classification decisions with deep taylor
  decomposition.
\newblock \emph{Pattern Recognition}, 65:211--222.

\bibitem[{Mou et~al.(2015)Mou, Men, Li, Xu, Zhang, Yan, and
  Jin}]{mou2015natural}
Lili Mou, Rui Men, Ge~Li, Yan Xu, Lu~Zhang, Rui Yan, and Zhi Jin. 2015.
\newblock Natural language inference by tree-based convolution and heuristic
  matching.
\newblock \emph{arXiv preprint arXiv:1512.08422}.

\bibitem[{Murdoch et~al.(2018)Murdoch, Liu, and Yu}]{murdoch2018beyond}
W~James Murdoch, Peter~J Liu, and Bin Yu. 2018.
\newblock Beyond word importance: Contextual decomposition to extract
  interactions from lstms.
\newblock \emph{arXiv preprint arXiv:1801.05453}.

\bibitem[{Pilault et~al.(2020)Pilault, Elhattami, and
  Pal}]{pilault2020conditionally}
Jonathan Pilault, Amine Elhattami, and Christopher Pal. 2020.
\newblock Conditionally adaptive multi-task learning: Improving transfer
  learning in nlp using fewer parameters \& less data.

\bibitem[{Rajani et~al.(2019)Rajani, McCann, Xiong, and
  Socher}]{rajani2019explain}
Nazneen~Fatema Rajani, Bryan McCann, Caiming Xiong, and Richard Socher. 2019.
\newblock Explain yourself! leveraging language models for commonsense
  reasoning.
\newblock \emph{arXiv preprint arXiv:1906.02361}.

\bibitem[{Ren et~al.(2019)Ren, Deng, He, and Che}]{ren-etal-2019-generating}
Shuhuai Ren, Yihe Deng, Kun He, and Wanxiang Che. 2019.
\newblock Generating natural language adversarial examples through probability
  weighted word saliency.
\newblock In \emph{Proceedings of the 57th Annual Meeting of the Association
  for Computational Linguistics}, pages 1085--1097, Florence, Italy.
  Association for Computational Linguistics.

\bibitem[{Ribeiro et~al.(2016)Ribeiro, Singh, and Guestrin}]{ribeiro2016i}
Marco~Tulio Ribeiro, Sameer Singh, and Carlos Guestrin. 2016.
\newblock "why should i trust you?": Explaining the predictions of any
  classifier.

\bibitem[{Rogers et~al.(2020)Rogers, Kovaleva, and
  Rumshisky}]{rogers2020primer}
Anna Rogers, Olga Kovaleva, and Anna Rumshisky. 2020.
\newblock A primer in bertology: What we know about how bert works.

\bibitem[{Samanta and Mehta(2017)}]{samanta2017towards}
Suranjana Samanta and Sameep Mehta. 2017.
\newblock Towards crafting text adversarial samples.
\newblock \emph{arXiv preprint arXiv:1707.02812}.

\bibitem[{Schwab and Karlen(2019)}]{schwab2019cxplain}
Patrick Schwab and Walter Karlen. 2019.
\newblock Cxplain: Causal explanations for model interpretation under
  uncertainty.
\newblock In \emph{Advances in Neural Information Processing Systems}, pages
  10220--10230.

\bibitem[{Selvaraju et~al.(2017)Selvaraju, Cogswell, Das, Vedantam, Parikh, and
  Batra}]{selvaraju2017grad}
Ramprasaath~R Selvaraju, Michael Cogswell, Abhishek Das, Ramakrishna Vedantam,
  Devi Parikh, and Dhruv Batra. 2017.
\newblock Grad-cam: Visual explanations from deep networks via gradient-based
  localization.
\newblock In \emph{Proceedings of the IEEE international conference on computer
  vision}, pages 618--626.

\bibitem[{Sennrich et~al.(2016)Sennrich, Haddow, and
  Birch}]{sennrich2016back-translation}
Rico Sennrich, Barry Haddow, and Alexandra Birch. 2016.
\newblock Improving neural machine translation models with monolingual data.
\newblock In \emph{Proceedings of the 54th Annual Meeting of the Association
  for Computational Linguistics (Volume 1: Long Papers)}, pages 86--96, Berlin,
  Germany. Association for Computational Linguistics.

\bibitem[{Seo et~al.(2016)Seo, Kembhavi, Farhadi, and
  Hajishirzi}]{seo2016bidirectional}
Minjoon Seo, Aniruddha Kembhavi, Ali Farhadi, and Hannaneh Hajishirzi. 2016.
\newblock Bidirectional attention flow for machine comprehension.
\newblock \emph{arXiv preprint arXiv:1611.01603}.

\bibitem[{Serrano and Smith(2019)}]{serrano2019attention}
Sofia Serrano and Noah~A Smith. 2019.
\newblock Is attention interpretable?
\newblock \emph{arXiv preprint arXiv:1906.03731}.

\bibitem[{Shi et~al.(2016)Shi, Knight, and Yuret}]{shi-etal-2016-neural}
Xing Shi, Kevin Knight, and Deniz Yuret. 2016.
\newblock Why neural translations are the right length.
\newblock In \emph{Proceedings of the 2016 Conference on Empirical Methods in
  Natural Language Processing}, pages 2278--2282, Austin, Texas. Association
  for Computational Linguistics.

\bibitem[{Shrikumar et~al.(2017)Shrikumar, Greenside, and
  Kundaje}]{shrikumar2017learning}
Avanti Shrikumar, Peyton Greenside, and Anshul Kundaje. 2017.
\newblock Learning important features through propagating activation
  differences.
\newblock \emph{arXiv preprint arXiv:1704.02685}.

\bibitem[{Simonyan et~al.(2013)Simonyan, Vedaldi, and
  Zisserman}]{simonyan2013deep}
Karen Simonyan, Andrea Vedaldi, and Andrew Zisserman. 2013.
\newblock Deep inside convolutional networks: Visualising image classification
  models and saliency maps.
\newblock \emph{arXiv preprint arXiv:1312.6034}.

\bibitem[{Socher et~al.(2013)Socher, Perelygin, Wu, Chuang, Manning, Ng, and
  Potts}]{socher2013recursive}
Richard Socher, Alex Perelygin, Jean Wu, Jason Chuang, Christopher~D Manning,
  Andrew~Y Ng, and Christopher Potts. 2013.
\newblock Recursive deep models for semantic compositionality over a sentiment
  treebank.
\newblock In \emph{Proceedings of the 2013 conference on empirical methods in
  natural language processing}, pages 1631--1642.

\bibitem[{Springenberg et~al.(2014)Springenberg, Dosovitskiy, Brox, and
  Riedmiller}]{springenberg2014striving}
Jost~Tobias Springenberg, Alexey Dosovitskiy, Thomas Brox, and Martin
  Riedmiller. 2014.
\newblock Striving for simplicity: The all convolutional net.
\newblock \emph{arXiv preprint arXiv:1412.6806}.

\bibitem[{Srinivas and Fleuret(2019)}]{srinivas2019full}
Suraj Srinivas and Fran{\c{c}}ois Fleuret. 2019.
\newblock Full-gradient representation for neural network visualization.
\newblock In \emph{Advances in Neural Information Processing Systems}, pages
  4124--4133.

\bibitem[{Strobelt et~al.(2016)Strobelt, Gehrmann, Huber, Pfister, Rush
  et~al.}]{strobelt2016visual}
Hendrik Strobelt, Sebastian Gehrmann, Bernd Huber, Hanspeter Pfister,
  Alexander~M Rush, et~al. 2016.
\newblock Visual analysis of hidden state dynamics in recurrent neural
  networks.
\newblock \emph{arXiv preprint arXiv:1606.07461}.

\bibitem[{Tenney et~al.(2019)Tenney, Das, and Pavlick}]{tenney2019bert}
Ian Tenney, Dipanjan Das, and Ellie Pavlick. 2019.
\newblock Bert rediscovers the classical nlp pipeline.
\newblock \emph{arXiv preprint arXiv:1905.05950}.

\bibitem[{Vashishth et~al.(2019)Vashishth, Upadhyay, Tomar, and
  Faruqui}]{vashishth2019attention}
Shikhar Vashishth, Shyam Upadhyay, Gaurav~Singh Tomar, and Manaal Faruqui.
  2019.
\newblock Attention interpretability across nlp tasks.
\newblock \emph{arXiv preprint arXiv:1909.11218}.

\bibitem[{Vaswani et~al.(2017)Vaswani, Shazeer, Parmar, Uszkoreit, Jones,
  Gomez, Kaiser, and Polosukhin}]{vaswani2017attention}
Ashish Vaswani, Noam Shazeer, Niki Parmar, Jakob Uszkoreit, Llion Jones,
  Aidan~N Gomez, {\L}ukasz Kaiser, and Illia Polosukhin. 2017.
\newblock Attention is all you need.
\newblock In \emph{Advances in neural information processing systems}, pages
  5998--6008.

\bibitem[{Vig and Belinkov(2019)}]{vig2019analyzing}
Jesse Vig and Yonatan Belinkov. 2019.
\newblock Analyzing the structure of attention in a transformer language model.
\newblock \emph{arXiv preprint arXiv:1906.04284}.

\bibitem[{Wiegreffe and Pinter(2019)}]{wiegreffe-pinter-2019-attention}
Sarah Wiegreffe and Yuval Pinter. 2019.
\newblock Attention is not not explanation.
\newblock In \emph{Proceedings of the 2019 Conference on Empirical Methods in
  Natural Language Processing and the 9th International Joint Conference on
  Natural Language Processing (EMNLP-IJCNLP)}, pages 11--20, Hong Kong, China.
  Association for Computational Linguistics.

\bibitem[{Williams(1992)}]{DBLP:journals/ml/Williams92}
Ronald~J. Williams. 1992.
\newblock Simple statistical gradient-following algorithms for connectionist
  reinforcement learning.
\newblock \emph{Machine Learning}, 8:229--256.

\bibitem[{Wu et~al.(2019)Wu, Wang, Yuan, Wu, and Li}]{wu2019coreference}
Wei Wu, Fei Wang, Arianna Yuan, Fei Wu, and Jiwei Li. 2019.
\newblock Coreference resolution as query-based span prediction.
\newblock \emph{arXiv preprint arXiv:1911.01746}.

\bibitem[{Yin et~al.(2020)Yin, Meng, and Chang}]{yin2020sentibert}
Da~Yin, Tao Meng, and Kai-Wei Chang. 2020.
\newblock Sentibert: A transferable transformer-based architecture for
  compositional sentiment semantics.

\bibitem[{Zeiler and Fergus(2014)}]{zeiler2014visualizing}
Matthew~D Zeiler and Rob Fergus. 2014.
\newblock Visualizing and understanding convolutional networks.
\newblock In \emph{European conference on computer vision}, pages 818--833.
  Springer.

\bibitem[{Zhang et~al.(2020{\natexlab{a}})Zhang, Sheng, Alhazmi, and
  Li}]{zhang2020adversarial}
Wei~Emma Zhang, Quan~Z Sheng, Ahoud Alhazmi, and Chenliang Li.
  2020{\natexlab{a}}.
\newblock Adversarial attacks on deep-learning models in natural language
  processing: A survey.
\newblock \emph{ACM Transactions on Intelligent Systems and Technology (TIST)},
  11(3):1--41.

\bibitem[{Zhang et~al.(2019)Zhang, Wu, Li, and Zhao}]{zhang2019explicit}
Zhuosheng Zhang, Yuwei Wu, Zuchao Li, and Hai Zhao. 2019.
\newblock Explicit contextual semantics for text comprehension.
\newblock In \emph{Proceedings of the 33rd Pacific Asia Conference on Language,
  Information and Computation (PACLIC 33)}.

\bibitem[{Zhang et~al.(2020{\natexlab{b}})Zhang, Wu, Zhao, Li, Zhang, Zhou, and
  Zhou}]{zhang2020semantics}
Zhuosheng Zhang, Yuwei Wu, Hai Zhao, Zuchao Li, Shuailiang Zhang, Xi~Zhou, and
  Xiang Zhou. 2020{\natexlab{b}}.
\newblock Semantics-aware bert for language understanding.
\newblock In \emph{Proceedings of the AAAI Conference on Artificial
  Intelligence}, volume~34, pages 9628--9635.

\end{thebibliography}
\bibliographystyle{acl_natbib}

\appendix

\section{Datasets and Training Details}
The Stanford Sentiment Treebank (SST) \citep{socher2013recursive} is a widely used benchmark for text classification. The task is to perform both fine-grained (very positive, positive, neutral, negative and very negative) and coarse-grained (positive and negative) sentiment classification at both the phrase and sentence level. We adopt the fine-grained setup in this work. We use Adam \citep{kingma2014adam} to optimize all modesl, with $\beta_1=0.9,\beta_2=0.98,\epsilon=10^{-8}$. All hyperparameters including batch size, dropout and learning rate are tuned on the validation set.

The Stanford Natural Language Inference (SNLI) Corpus \citep{bowman2015large} is a collection of 570k human-written English sentence pairs for the task of natural language inference (NLI). It contains a balanced amount of sentence pairs with  labels entailment, contradiction, and neutral. We follow the official train/dev/test splits for both datasets. We also use Adam \citep{kingma2014adam} for optimization. 
Hyperparameters including batch size, dropout and learning rate are tuned on the validation set 

%to optimize all modesl, with $\beta_1=0.9,\beta_2=0.98,\epsilon=10^{-9}$. All .

IWSLT2014 En$\rightarrow$De contains 160k training sequences pairs and 7k validation sentence pairs. We use the concatenation of dev2010, tst2010, tst2011 and tst2011 as the test set. A joint BPE vocabulary of about 10k tokens is constructed.  
We minimize the cross entropy loss with label smoothing of value 0.1 during training.  Adam \citep{kingma2014adam} is used for optimization.

IMDB contains an even number of positive and negative reviews. The training and test sets  respectively contain 25k and 25k examples.
Yahoo! Answers is a large dataset for topic classification over ten largest main categories from Yahoo! Answers Comprehensive Questions and Answers v1.0. This dataset contains 1,400k training exmaples and 60k test examples respectively.

\end{document}